\documentclass[authoryear,review,3p,times]{elsarticle}

\usepackage{amssymb}
\usepackage{amsmath}
\usepackage{lineno}
\usepackage{bm} 
\usepackage{longtable}
\usepackage{caption}
\usepackage{soul}
\usepackage{color}
\usepackage{gensymb}
\usepackage[flushleft]{threeparttable}
\usepackage{algorithm}
\usepackage{algpseudocode}
\usepackage{hyperref}
\usepackage{nomencl} 
\makenomenclature

\graphicspath{{figures/}}

\usepackage{array}

\journal{Nuclear Engineering and Design}

\begin{document}

\begin{frontmatter}



\title{Techno-economic optimization of a heat-pipe microreactor, part II: multi-objective optimization analysis}


\author[1]{Paul Seurin \corref{cor1}} 
\ead{paul.seurin@inl.gov}
\author[2]{Dean Price} 

\affiliation[1]{organization={Autonomous Engineering Department, Idaho National Laboratory}, 
            addressline={1955 Fremont Ave.}, 
            city={Idaho Falls},
            postcode={83402}, 
            state={ID},
            country={US}}

\affiliation[2]{organization=Massachusetts Institute of Technology, 
            addressline={77 Massachusetts Ave.}, 
            city={Cambridge},
            postcode={02935}, 
            state={MA},
            country={US}}

\begin{abstract}
Heat-pipe microreactors (HPMRs) are compact and transportable nuclear power systems exhibiting inherent safety, well-suited for deployment in remote regions where access is limited and reliance on costly fossil fuels is prevalent. In prior work, we developed a design optimization framework that incorporates techno-economic considerations through surrogate modeling and reinforcement learning (RL)-based optimization, focusing solely on minimizing the levelized cost of electricity (LCOE) by using a bottom-up cost estimation approach. In this study, we extend that framework to a multi-objective optimization that uses the Pareto Envelope Augmented with Reinforcement Learning (PEARL) algorithm. The objectives include minimizing both the rod-integrated peaking factor ($F_{\Delta h}$) and LCOE---subject to safety and operational constraints. We evaluate three cost scenarios: (1) a high-cost axial and drum reflectors, (2) a low-cost axial reflector, and (3) low-cost axial and drum reflectors. Our findings indicate that reducing the solid moderator radius, pin pitch, and drum coating angle---all while increasing the fuel height---effectively lowers $F_{\Delta h}$. Across all three scenarios, four key strategies consistently emerged for optimizing LCOE: (1) minimizing the axial reflector contribution when costly, (2) reducing control drum reliance, (3) substituting expensive tri-structural isotropic (TRISO) fuel with axial reflector material priced at the level of graphite, and (4) maximizing fuel burnup. While PEARL demonstrates promise in navigating trade-offs across diverse design scenarios, discrepancies between surrogate model predictions and full-order simulations remain. Further improvements are anticipated through constraint relaxation and surrogate development, constituting an ongoing area of investigation.
\end{abstract}



\end{frontmatter}

\nomenclature{QoI}{quantity of interest}
\nomenclature{SDM}{shutdown margin}
\nomenclature{$F_{\Delta h}$}{rod-integrated peaking factor}
\nomenclature{$q^{''}_{max}$}{peak heat flux}
\nomenclature{ITC}{isothermal temperature coefficient}
\nomenclature{TC}{temperature coefficient}


\nomenclature{AI}{artificial intelligence}

\nomenclature{PEARL}{Pareto Envelope Augmented with Reinforcement Learning}
\nomenclature{RL}{reinforcement learning}
\nomenclature{PPO}{proximal policy optimization}
\nomenclature{MOO}{multi-objective optimization}
\nomenclature{$\mu$R}{microreactor}
\nomenclature{HPMR}{heat-pipe microreactor}
\nomenclature{HP}{heat pipe}
\nomenclature{TRISO}{tri-structural isotropic}

\nomenclature{LCOE}{levelized cost of electricity}
\nomenclature{O\&M}{operation and maintenance}
\nomenclature{FOAK}{first of a kind}
\nomenclature{MOUSE}{Microreactor Optimization Using Simulation and Economics}

\nomenclature{CPU}{central processing unit}
\nomenclature{NSGA-II}{Non-dominated Sorting Genetic Algorithm}

\printnomenclature


\newcolumntype{L}[1]{>{\raggedright\arraybackslash}p{#1}}
\renewcommand{\hl}[1]{#1}

\section{Introduction}
\label{sec:intro}
Microreactors ($\mu$Rs), defined by the U.S.~Department of Energy as nuclear reactors that produce less than 20 MWth, fall under Category B reactors and are classified as Category 2 hazards per 10 CFR 830 \citep{naranjo2024assessment}. Designed for factory assembly, transportability (via truck or rail), simplified installation, and autonomous operation, $\mu$Rs are particularly well-suited for deployment in remote or unmanned locations where reliance on costly fossil fuels remains prevalent (e.g., deep-sea platforms, industrial mining sites, and military installations) \citep{wang2025data, buongiorno2021can}. Despite their promise, $\mu$Rs face diseconomies of scale and must leverage economies of multiples and favorable clean energy incentives to achieve market competitiveness \citep{abdusammi2025evaluation}. While benefits such as standardization, passive safety, reduced radionuclide inventories, and lower financing costs are anticipated \citep{abou2021economics}, these advantages have yet to be fully realized.
 
Within the broader class of $\mu$Rs, heat-pipe microreactors (HPMRs) utilize heat pipes (HPs) for passive heat removal, eliminating the need for active circulation systems and enabling highly compact designs. HPs, widely used in electronics as isothermal heat exchangers, offer a promising pathway to portable and potentially cost-effective nuclear systems \citep{yan2020technology}. Westinghouse’s eVinci$^{TM}$ program exemplifies ongoing efforts to commercialize HPMRs \citep{price2024multiphysics}. However, HPMRs often struggle to compete economically with alternative designs such as gas-cooled or lead-cooled $\mu$Rs \citep{hanna2025bottom}, underscoring the need for targeted economic optimization strategies.
 
Previous efforts to improve $\mu$Rs economics have focused on fixed geometric configurations, market analyses \citep{buongiorno2021can, park2025bottom}, fuel cycle cost reduction \citep{al2025assessing}, material selection \citep{shirvan2023uo2}, automation \citep{naranjo2024assessment}, power uprates \citep{park2025bottom}, and multi-unit deployment strategies \citep{bryan2023remote}. Concurrently, geometric design optimization has been explored by using multiphysics simulations to enhance power density and fuel enrichment \citep{wang2025data} or to minimize thermal stress for structural integrity \citep{zhang2025optimizing}, often employing the Non-dominated Sorting Genetic Algorithm (NSGA-II) \citep{deb2002fast}. These approaches are challenged by the complex coupling of neutronic, thermal, and mechanical phenomena in HPMRs, and thus require computationally intensive simulations (e.g., OpenMC \citep{romano2015openmc}) and surrogate modeling in order to remain tractable. In our companion study \citep{seurin2025techno}, we introduced a unified optimization framework that evaluates the impact of geometric design decisions on the economic performance of HPMRs. By varying the cost of the axial reflectors (from \$45,000/kg to \$80/kg [2024 USD]) and imposing safety and operational constraints (including peak heat flux $q^{''}_{max}$, rod-integrated peaking factor $F{\Delta h}$, shutdown margin (SDM), and fuel lifetime), we demonstrated that the levelized cost of electricity (LCOE) for a first-of-a-kind (FOAK) HPMR could be reduced by up to 57\%. However, these results were sensitive to cost assumptions and conservative safety constraints.

A key limitation in HPMR design is the lack of understanding regarding the trade-offs between safety, performance, and economic metrics---particularly for reactor technologies that have yet to be built. Designers often rely on iterative, resource-intensive processes to tailor systems to specific applications, and the design space remains underexplored. Thus, there is a pressing need for tools that can holistically and efficiently navigate these trade-offs, enabling more granular and proportionate design adjustments.
Recent advances in open-source tools (e.g., Microreactor Optimization Using Simulation and Economics (MOUSE) for techno-economic modeling \citep{hanna2025bottom}, OpenMC for Monte Carlo simulations \citep{romano2015openmc}, scikit-learn for surrogate modeling \citep{scikit-learn}, and the Pareto Envelope Augmented with Reinforcement Learning (PEARL) algorithm for multi-objective optimization (MOO) \citep{seurin2024multiobjective}) offer a promising foundation for such capabilities.
Importantly, the challenge addressed here is emblematic of broader trends in nuclear engineering, a field in which MOO is increasingly essential. The field integrates diverse disciplines, each with distinct goals and constraints, thus requiring reconciliation of conflicting objectives such as cost, safety, regulatory compliance, and operational performance. Nuclear engineering has benefited from a range of MOO algorithms, including NSGA-II/III \citep{parks1996multiobjective}, Pareto Ant Colony Optimization, the Multi-Objective Cross-Entropy Method \citep{schlunz2016acomparative}, Multi-Objective Tabu Search \citep{mawdsley2022incore}, reinforcement learning (RL) \citep{seurin2024multiobjective}, and Multi-Objective Simulated Annealing \citep{park2009multiobjective}. Applications span core loading pattern optimization \citep{kropaczek2009copernicus}, fuel performance \citep{seurin2025impact}, and power conversion system design \citep{song2024multi, raiyan2025exergoeconomic}.

However, these techniques have not been extensively adopted in the nuclear sector, unlike fields (e.g., chemical engineering \citep{stewart2021asurvey} or thermodynamics of heat engines \citep{kumar2016multiobjective}) in which reliance on these tools has grown thanks to advancements in computational power. More specifically, recent advances in artificial intelligence (AI), particularly RL \citep{seurin2024multiobjective}, have demonstrated the potential to efficiently and rapidly characterize safety and economic performance trade-offs in the design of large-scale light-water reactors. While design considerations vary depending on the application, this prior work showed RL algorithms to be broadly applicable across different reactor types and design objectives. Moreover, AI-based algorithms can be enhanced by incorporating engineering intuition into their structure \citep{seurin2024physics}---particularly by accounting for correlations between physical parameters that couple constraints with objectives, thus potentially yielding safer, more economical designs.
The objective of this work is not to present a finalized optimized design for the HPMR model, as both the reference design (described in Section \ref{sec:nomdes}) and the techno-economic analysis of $\mu$Rs remain subject to significant uncertainties. Instead, the goal is to obtain human-understandable insights into the safety and cost trade-offs that emerge from applying MOO to geometric design decisions. These insights represent the novel contribution of this study, and are emphasized over specific performance gains.
Three cost scenarios are considered, each leading to distinct optimal design strategies: 

\begin{enumerate}
    \item Scenario 1: Expensive Beryllium Axial Reflector (Compact Design).  The beryllium offers superior moderation for compactness but at a high cost. 
\item Scenario 2: Graphite Axial Reflector (Cost-Optimized). Shift to graphite in the axial reflector for cost savings, sacrificing some compactness.
\item Scenario 3: Graphite Axial and Drum Reflectors (Fully Cost-Optimized). Both axial and drum reflectors use graphite for maximum cost reduction, for the least compact design.
\end{enumerate}

The contributions of this work are summarized as follows:
\begin{itemize}
    \item FOAK MOO of HPMRs, incorporating techno-economic considerations
    \item Knowledge synthesis through characterization of performance trade-offs in $\mu$R design, balancing safety and cost
    \item Identification of key cost-driver trade-offs with safety across different scenarios, as enabled by recent advances in RL-based optimization, specifically PEARL.
\end{itemize}

The remainder of the paper is organized as follows. In Section \ref{sec:methodology}, we present the nominal design upon which we will base our optimization (see Section \ref{sec:nomdes}), the physics and techno-economic modeling underlying our candidate evaluation (see Section \ref{sec:physicsandtechnoecon}), and the optimization methodology, including the formulation of the objective function and PEARL (see Section \ref{sec:optimiztionmethodoogy}). The results are then given in Section \ref{sec:res}, along with both a description of the data generated for the surrogate model and a correlation matrix (see Section \ref{sec:samplesdescription}), the application of PEARL with the analysis of the best designs found (see Section \ref{sec:applicationof}), and a summary of the results obtained (see Section \ref{sec:summary}). Lastly, concluding remarks and areas of future research are given in Section \ref{sec:conc}.

\section{Methodology}
\label{sec:methodology}

In \citep{seurin2025techno}, a detailed description of the physics and the techno-economic models used in the current work is provided. For the sake of brevity and to avoid repetition, we invite readers to explore that paper to understand the full modeling framework. Here, we only briefly recall those key factors that influenced the design decisions.

\subsection{Nominal Reactor Design, Design Parameterization, and Bounds}
\label{sec:nomdes}

The HPMR design utilized in this study was derived from a modeling experiment aimed at showcasing advanced modeling and simulation capabilities \citep{stauff2022multiphysics}. Although the design has been modified since its initial conception, it remains a valuable candidate for the current methodology thanks to its flexibility and potential for improvement. Notably, the original design was developed without consideration of real-world implementation constraints, allowing for broader exploration of optimization strategies.

\begin{figure}[htb!]
    \centering
    \includegraphics[width=.9\textwidth]{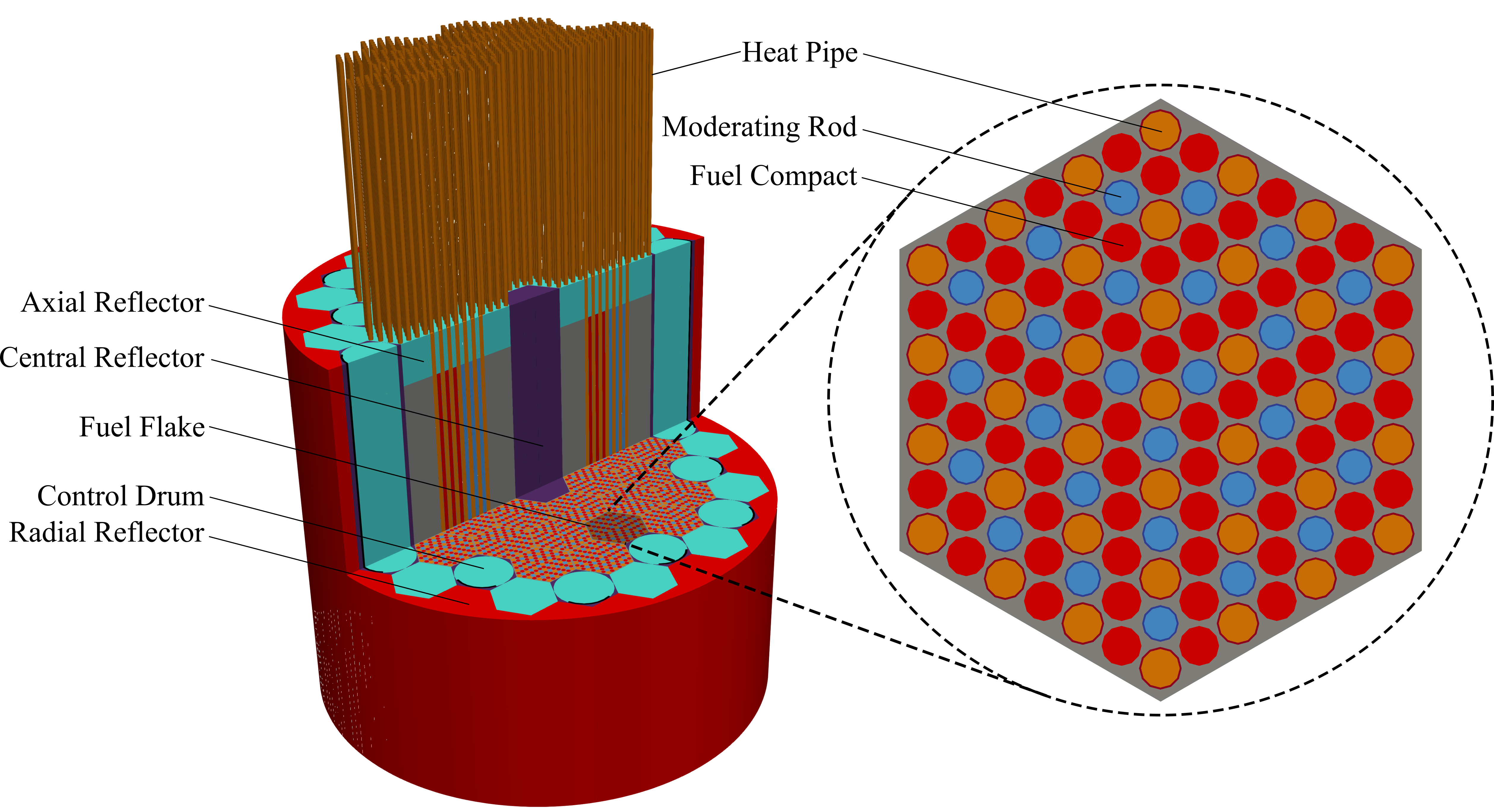}
    \caption{Nominal reactor design with labeled components.}
    \label{fig:rctr_diag}
\end{figure}

Figure \ref{fig:rctr_diag} outlines the key design characteristics of the nominal configuration of the HPMR analyzed in this study. Table \ref{tab:inparms} presents the seven design parameters that were focused on in this study, along with their nominal values, bounds, and corresponding mathematical symbols. The fuel compact radius and moderator radius were constrained by the pin pitch so as to ensure geometric feasibility of the $\mu$R core layout. The remaining bounds were selected in order to encompass a design space that includes configurations relevant to potential optimal solutions.

\begin{table}[htb!]
\centering
\caption{Parameter descriptions for changing aspects of candidate designs.}
\label{tab:inparms}
\begin{tabular}{llll}
\hline
Symbol    & Description                           & Nominal Value & Bounds                                                                                                     \\ \hline
$x_{ca}$  & Control drum coating angle            & 90$^\circ$     & [35, 180]                                                                                                  \\
$x_{B10}$ & Control drum absorber B-10 enrichment & 95\%         & [20\%, 95\%]                                                                                               \\
$x_{fh}$  & Active fuel height                    & 160 cm        & [130 cm, 190 cm]                                                                                           \\
$x_{pp}$  & Pin pitch                             & 2.3 cm        & [1.94 cm, 2.78 cm]                                                                                         \\
$x_{e}$   & U-235 fuel enrichment                 & 19.7\%        & [17\%, 20\%]                                                                                             \\
$x_{cr}$  & Fuel compact radius                   & 1 cm          & $\left[\frac{1}{4}x_{pp}, \frac{1}{2}x_{pp}\right]$                                                                 \\
$x_{mr}$  & Moderator radius                      & 0.825 cm      & $\left[\frac{1}{5}\left(x_{pp}-2(0.095 \text{ cm})\right), \frac{1}{2}\left(x_{pp}-2(0.095 \text{ cm})\right)\right]$ \\ \hline
\end{tabular}
\end{table}

\subsection{Physics and Techno-economic Modeling}
\label{sec:physicsandtechnoecon}

The two major modeling components of this work are physics modeling and techno-economic modeling. Physics modeling is essential for estimating quantities of interest (QoIs), while techno-economic modeling enables financial cost-benefit analyses of $\mu$R designs.

For this study, the OpenMC \citep{romano2015openmc} Monte Carlo code was employed to simulate candidate reactor designs. OpenMC, a continuous-energy neutron transport code that utilizes the ENDF/B-VIII.0 cross-section libraries \citep{brown2018endf}, was used to extract key QoIs related to cost and neutronics (e.g., fuel lifetime, SDM, temperature coefficients (TCs), and peaking factors).

\begin{enumerate}
    \item \textbf{Lifetime}: The operational lifetime or fuel lifetime significantly impacts the economic value proposition of the $\mu$R.
    \item \textbf{SDM}: Expressed in pcm, the SDM quantifies the available reactivity margin for safely shutting down the reactor from operating conditions. Its calculation is detailed in \citep{seurin2025techno}.
    \item \textbf{TCs}: To ensure inherent safety, nuclear reactors must exhibit negative TCs, such that an increase in temperature leads to a reduction in reactivity. In HPMRs, competing effects include Doppler broadening (i.e., increased neutron absorption cross sections) and temperature feedback from HP walls (i.e., changes in capture cross sections). The isothermal TC (ITC) was evaluated as a worst-case scenario. The method for its calculation is provided in \citep{seurin2025techno}.
    \item \textbf{Peaking Factors}: As proxies for power and temperature-induced effects, two peaking factors were considered in this study: $F_{\Delta h}$, the maximum rod-integrated peaking factor \citep{seurin2024assessment}, and the peak heat flux, $q^{''}_{\text{max}}$. The product of $q^{''}_{\text{max}}$ and $F_{\Delta h}$ serves as an indicator of the maximum temperatures experienced by the fuel, monolith, reflector, solid moderator, and HP, as well as the peak energy transferred to the HP. Their respective calculation methods are described in \citep{seurin2025techno}.
\end{enumerate}

The cost of a $\mu$R can be assessed by using bottom-up cost estimation tools \citep{park2025bottom,al2025open} in conjunction with the Code of Accounts methodology. This approach calculates the cost of individual components, including fuel, operation and maintenance (O\&M) costs (e.g., labor, spares, taxes, and fees), and direct and indirect capital costs. The MOUSE tool---currently under development at Idaho National Laboratory---was employed \citep{hanna2025bottom}. Baseline cost assumptions are provided in the associated references. The LCOE was computed by summing the discounted costs of fuel, O\&M, and capital over the reactor's lifetime, then dividing by the discounted energy produced over that same period:
\begin{equation}
    LCOE = \frac{\sum_{t=0}^n\frac{F_t+O_t+TCI_t}{(1+r)^t}}{\sum_{t=0}^n\frac{E_0}{(1+r)^t}}
\end{equation}
where $F_t$ is the cost of fuel in year t; $O_t$ is the O\&M cost in year t; $TCI_t$ is the total capital invested, representing the sum of the overnight capital cost and financial cost (e.g., interest, escalation, and contingencies) in year t; $E_t$ is the generated electrical energy in year t; r is the discount rate; and n is the expected plant life in years. r is assumed to be 6\%, while n is assumed to be 60 years.
The QoIs associated with the nominal design will be presented when comparing against the MOO's solution.

\subsection{Optimization Methodology: PEARL}
\label{sec:optimiztionmethodoogy}

PEARL \citep{seurin2024multiobjective}, a MOO algorithm based on RL, is designed to address challenges in operations research, engineering, business, and other domains that inherently involve multiple competing objectives. It supports both minimization and maximization tasks, accommodates constrained and unconstrained optimization, and handles both integer and continuous input spaces. PEARL was developed to overcome the limitations of classical heuristic-based approaches, which typically rely on stochastic search strategies and rule-of-thumb heuristics. Compared to multi-objective variants of Tabu Search, Simulated Annealing, and Genetic Algorithms (e.g., NSGA-II \citep{deb2002fast} and NSGA-III \citep{deb2013evolutionary,jain2013evolutionary}), PEARL has demonstrated superior performance across several metrics---including hypervolume and the number of feasible solutions---on both benchmark mathematical problems and large-scale pressurized-water reactor optimization tasks. This advantage stems from RL’s learning mechanism, previously explored in single-objective RL frameworks \citep{seurin2024assessment,seurin2022pwr,seurin2025surpassing,seurin2023can}.

Unlike traditional methods that rely on random sampling to explore the design space, RL progressively learns which solutions to generate, based on feedback from the environment. This results in increasingly refined solutions over time, reducing reliance on randomness. The learning process can be further accelerated by incorporating domain-specific knowledge, referred to as ``engineering intuition'' in Section \ref{sec:intro} and in \citep{seurin2024physics}. Specifically, accounting for correlations between physical parameters that couple constraints with objectives can lead to safer, more economical designs. PEARL has been successfully applied to several nuclear engineering challenges, including optimizing fuel configurations in nuclear power plants \citep{seurin2024physics} and evaluating the feasibility of power uprates \citep{seurin2025impact}.

In the multi-objective context, PEARL evaluates solution quality---not through scalar aggregation (e.g., weighted sums), but via Pareto-front-based metrics such as crowding distance \citep{deb2002fast} or niching distance \citep{deb2013evolutionary}. During each rollout, a candidate solution is compared against a buffer of Pareto-optimal solutions. If the buffer has reached its maximum size $\kappa$, the new solution is ranked based on its distance metric and then inserted into the buffer, replacing the least favorable solution (rank $\kappa+1$). If the buffer is not yet full, all solutions are retained and re-arranged based on their new ranks (accounting for the new solution). This ranking serves as the reward signal for the RL agent, guiding the learning process toward generating Pareto-optimal solutions.

The concept of Pareto optimality generalizes scalar superiority to multi-objective contexts: a design is considered superior if it improves all objectives relative to another; otherwise, the designs are deemed equivalent. Departing from a Pareto-optimal solution necessarily degrades at least one objective \citep{seurin2024multiobjective}. Infeasible solutions are always dominated by feasible ones, and PEARL’s extension to constrained optimization is discussed in detail in \citep{seurin2024multiobjective}.
 The pseudocode for this procedure is summarized in Algorithm \ref{alg:PEARLepeusocode}:
\begin{algorithm}[H]
 \small
   \caption{Pseudocode for PEARL algorithms at time step t.}
   \label{alg:PEARLepeusocode}
   \begin{algorithmic}[1]
      \State \textbf{Input:}  maximum size of the buffer $\kappa$, buffer $E_b^{t - 1} = \{r_b^1,r_b^2,...,r_b^k\}$, learnable parameter at time t-1 $\theta_{t-1}$, type of distance metric $\mathcal{D}$
 \State \textbf{Output:}  Reward assigned to the new solution generated $\tilde{r_t}$ and Pareto front $E_b^{t}$ at time step t 
 \State \textbullet A solution $a_p$ is generated the current policy $a_p \sim \pi(\cdot;\theta_{t-1})$
                    \State \textbullet Solution $a_p$ is  evaluated by the environment and a reward $r_p \in \mathbb{R}^F$, where $F$ is the number of objectives, is returned.
                    \State \textbullet Apply a distance-based technique to obtain the rank of the solution generated in the buffer $E_b^{t-1}+\{r_p\}$ based on $\mathcal{D}$
                \State \textbullet Set $\tilde{r_t} = - \text{rank}(r_p)$
                \State \textbullet Obtain $\mathcal{PF}^t$ the Pareto front of $E_b^{t-1}+\{r_p\} = \{r_b^1,r_b^2,...,r_b^k,r_p\}$
                \State \textbullet Set $E_b^{t} = \mathcal{PF}^t[:\kappa]$
                \State \textbullet return $\tilde{r_t}$,$E_b^{t}$
\end{algorithmic}
\end{algorithm}

 Lastly, in \citep{seurin2024multiobjective}, proximal policy optimization (PPO) \citep{schulman2017proximal} was used; however, it should be noted that PPO was chosen here for its simplicity to tune and for its performance, but any other policy-based RL algorithm can be utilized and were tested, including trust region policy optimization \citep{schulman2017trust},  asynchronous advantage actor-critic \citep{mnih2016asynchronous}, and actor-critic using Kronecker-factored trust region \citep{schulman2017trust}.
 
\section{Results}
\label{sec:res}
\subsection{Samples Description and Objective Formulation}
\label{sec:samplesdescription}
We collected 921 samples by using the Sawtooth High-Performance Computing Center at Idaho National Laboratory, utilizing Intel(R) Xeon(R) Platinum 8268 central processing units (CPUs) clocked at 2.90 GHz per candidate. The computing time for each OpenMC candidate ranged from approximately 40 to 168 hours (i.e., the maximum allowable compute time on a Sawtooth node), depending on the fuel lifetime. The QoIs related to neutronics calculations were then extracted, and the fuel lifetime was input into the MOUSE techno-economic assessment tool to obtain the cost performance parameters.

Ultimately, we retained 875 data points and fitted a Gaussian process to predict fuel lifetime, SDM, and $F_{\Delta h}$, as well as a multi-layer perceptron for the $q^{''}_{max}$, while the mean LCOE cost was evaluated with a reduced number of stochastic samples (10).

\begin{figure}[htp!]
     \centering
\includegraphics[width=0.7\linewidth]{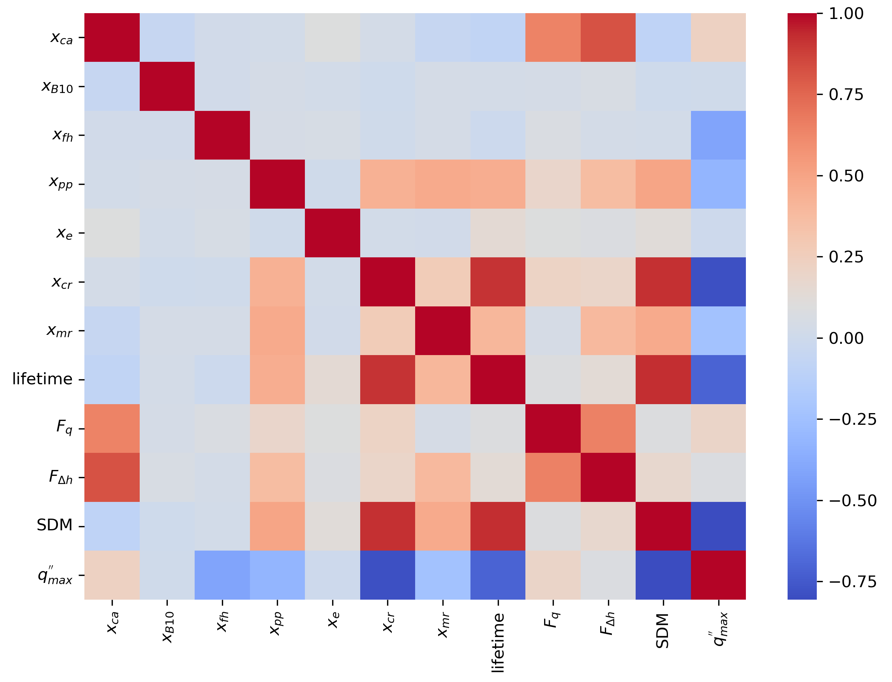}
     \caption{Correlation matrix between the input design parameters and QoIs. ``1'' signifies a strong positive linear correlation, and ``-1'' a strong negative one.}
     \label{fig:correlationmatrix}
 \end{figure}

A principle approach for finding the best combination of objectives and constraints is described in \citep{seurin2024physics}. The importance of selections between the constraints and the objectives is incumbent upon the physical relationship between each of the QoIs and design parameters. It was termed ``physics-informed PEARL'' in \citep{seurin2024physics} (and to a greater extent ``physics-informed optimization''). To understand the relationships, Figure \ref{fig:correlationmatrix} provides the (linear) correlation matrix between the input design parameters and the QoIs.
 
In this figure, we observe that both fuel lifetime and SDM are highly positively correlated with the compact radius $x_{cr}$. The compact radius serves as a proxy for the number of tri-structural isotropic (TRISO) in the core, which evidently enhances the reactivity in the reactor, thereby influencing both parameters. Surprisingly, enrichment shows a weaker correlation with these factors than would typically be expected in a large-scale reactor \citep{seurin2024assessment}. This indicates that the impact of the compact radius is more significant than the effect of increasing the $U_{235}$ content. However, this could be owed to the small. In other words, the net effect of increasing $x_{cr}$ while decreasing the enrichment is positive for the fuel lifetime and negative for the SDM. Note, however, that the enrichment range was limited to 3\% (see Table \ref{tab:inparms}), though the contribution would have certainly been higher had we been able to expand it. 

Secondary effects include the moderator radius $x_{mr}$ and pin pitch $x_{pp}$, both of which contribute to increased moderation. Interestingly, we had anticipated that the $B_{10}$ enrichment and coating angle $x_{ca}$ would strongly impact the SDM. However, since we perturbed all the design parameters simultaneously, only the strongest correlations were captured in this analysis.

$F_{\Delta h}$ is highly correlated with the coating angle of the control drums, and to a lesser extent with $x_{pp}$ and $x_{mr}$. It is important to note that the peaking factors were assessed when the drums were inserted at a 90-degree angle. A wider coating angle reduces leakage, thereby improving neutron economy; however, it also affects $F_{\Delta h}$ by redirecting more neutrons back to fission within the core asymmetrically. Similarly, $x_{mr}$ would increase moderation locally and reduce the mean free path of the neutrons, thereby increasing peaking.

The peak heat flux $q^{''}_{max}$, which is equal to $q^{''}_{avg} \times F_q$, is highly negatively correlated with both the compact radius $x_{cr}$ and the fuel height $x_{fh}$, as well as with the fuel lifetime and SDM. The negative correlation with $x_{cr}$ and $x_{fh}$ can be attributed to the geometry, as reflected in the formula for average heat flux: $q^{''}_{avg} = \frac{P}{N{flakes} \times N_{compacts} \times 2\pi x_{cr} x_{fh}}$. To a lesser extent, $x_{pp}$ and $x_{mr}$ are also negatively correlated to $q^{''}_{max}$. For the latter correlations, in addition to their common association with the two previous design parameters, the increase in reactivity within the core---often the result of localized higher moderation---plays a significant role similar to that of $F_{\Delta h}$.

For the application of PEARL, we chose to solve a constrained two-objective problem that seeks to balance the trade-offs between HPMR safety and cost efficiency. This includes optimizing the LCOE and $F_{\Delta h}$. 

 We have then established constraints for the $q^{''}_{max}$, SDM, fuel lifetime, and $F_{\Delta h}$; namely, 0.025 (to leave some leeway for cost optimization, as in some instances $q^{''}_{max}$ would conflict with optimizing the LCOE \citep{seurin2025techno}), -6700 pcm (close to the nominal design), 6.0 years (close to the nominal design), and 1.47 (the constraint imposed for the single-objective scenario in \citep{seurin2025techno}), respectively. As explained in Section \ref{sec:physicsandtechnoecon}, replacement of the main equipment occurs every 10 years, making it unnecessary to purchase additional fuel to exceed this lifetime. Therefore, we prescribed a fuel lifetime upper limit of 10.40 years so as to adopt a slightly conservative approach. A summary of the assignment of constraints and objectives is provided in Table \ref{tab:assignmentof}.

\begin{table}[htp!]
    \centering
\caption{Assignment of the objectives and constraints for the optimization with PEARL. $F_q$ is not utilized here, as it is only an input to evaluate the peak heat flux.}
    \begin{tabular}{lllll}
    \hline
    QoIs & Objectives & Constraints & Limits & Sign\\
    \hline
  LCOE & + & N/A & N/A & decrease \\
  $q^{''}_{max}$ & N/A & N/A & 0.025 & decrease \\ 
  $F_{\Delta h}$ & + & + & 1.47 & increase\\
  SDM [pcm] & N/A & + & -6700 & decrease\\
  Lifetime & N/A & + & [6.0, 10.40] & increase\\
  \hline
\end{tabular}
    \label{tab:assignmentof}
\end{table}


\subsection{Application of Optimization Methodology and Analysis of the Optimized Designs}
\label{sec:applicationof}

For this work, we used PEARL-niching, in which the distance criteria $\mathcal{D}$ is based on niching \citep{deb2013evolutionary}. The constraints were first solved and then the optimization in the multi-objective space (i.e., for all feasible solutions) was addressed. The constraints were evaluated as  $\sum_{i \in C} \gamma_i \Phi(x_i)$, where $C = \{ (c_{i})_{i \in \{1,..,4\}} \}$ is the set of constraints and $(\gamma_{i})_{i \in \{1,..,4\}}$ is a set of weights attributed to each constraint (all weights are equal to 10,000 in this work). If $ (x_{i})_{i \in \{1,..,4\}}$ are the values reached for the core ($x$) for the corresponding constraints $(c_{i})_{i \in \{1,..,4\}}$, $\Phi(x_i) = \delta_{x_i \le c_i} (\frac{x_i - c_i}{c_i})^2$, where $\delta$ is the Kronecker delta function.

The PPO-related hyperparameters that influence the optimization (see stables-baselines3 package \citep{stable-baselines3} and reference \citep{seurin2024assessment} for a description of their influence) are n$_{\text{steps}}$ equal to 8, a $c_{\mathcal{H}}$ of 0.0001, a $c_{vf}$ of 0.5, a learning rate of 0.00025, a maximum grad norm of 0.5, a batch size of 0.5, and a $\epsilon$ of 0.2. To recall, neither $\gamma$ nor $\lambda$ affect the optimization. For PEARL, only two hyperparameters are necessary: the $\kappa$ is 64 and the number of agents is eight. We then ran the PEARL algorithm \citep{seurin2024multiobjective} on eight 48 Intel(R) Xeon(R) Platinum 8268 CPUs clocked at 2.90 GHz, for 100,000 steps total. 

We examined three optimization scenarios corresponding to different material cost assumptions for the reflector and control drums, which were identified as major cost drivers in \citep{seurin2025techno}. While the previous study did not consider inexpensive control drums, in \ref{sec:applicationofbecostofgraphiteeveruwhere} we expand said analysis for the sake of completeness, providing insights into the safety and cost trade-offs for an additional $\mu$R cost assumption. Section \ref{sec:applicationofbecostofnominal} presents the nominal case in which the cost of beryllium (Be) is high, allowing us to draw key conclusions and observe the manifestation of physics-informed RL \citep{seurin2024physics}. 

\subsubsection{Scenario Analysis: Expensive Beryllium Axial and Control Drum Reflectors}
\label{sec:applicationofbecostofnominal}
\begin{figure}[htp!]
    \centering
    \includegraphics[width=0.5\linewidth]{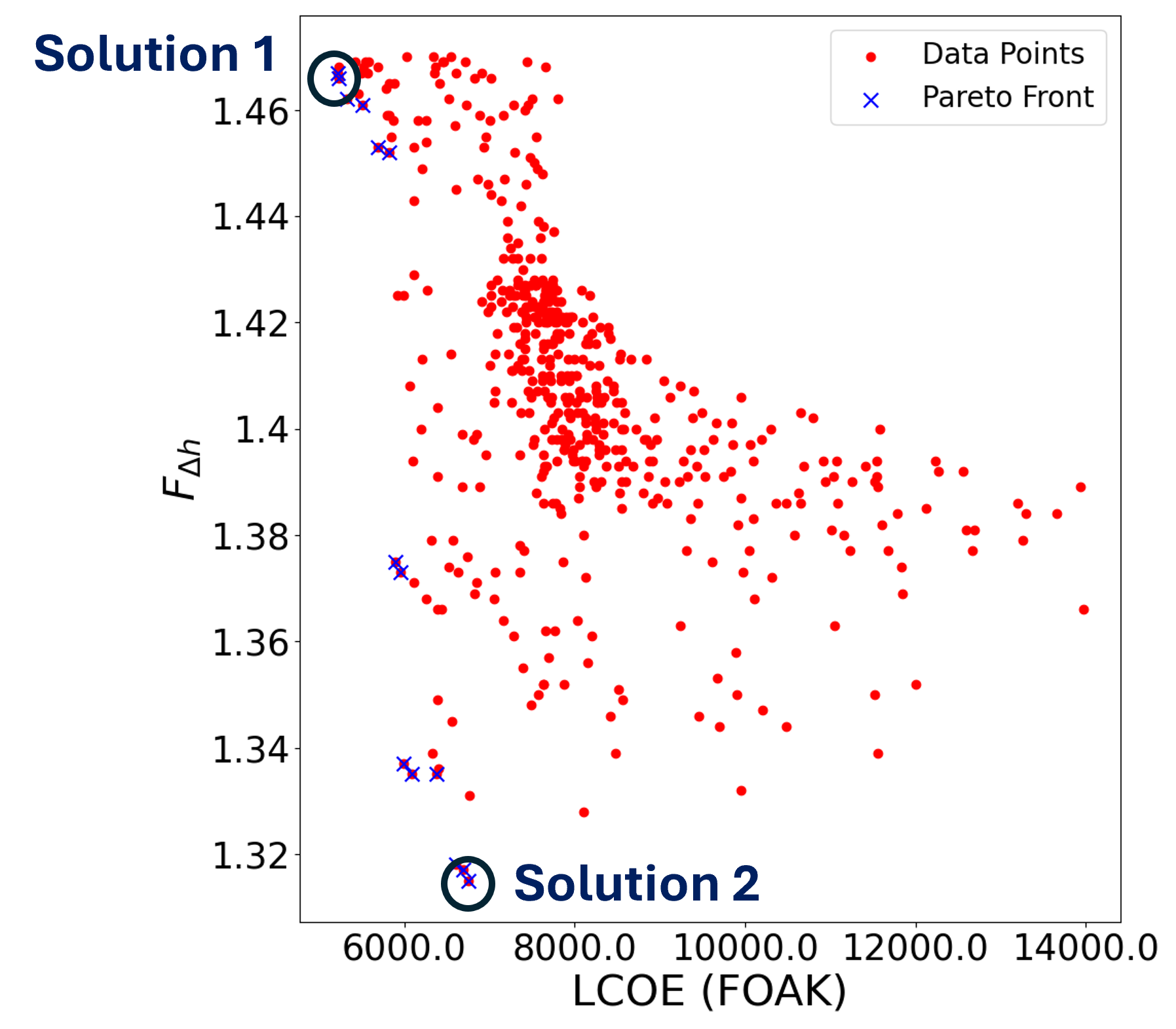}
    \caption{Pareto fronts from PEARL agents for the scenario involving expensive reflectors. The two solutions corresponding to the minimum LCOE (i.e., Solution 1) and $F_{\Delta h}$ (i.e., Solution 2) are circled in blue.}
    \label{fig:pearlresultsnominal}
\end{figure}

In this scenario, both the axial and the control drum reflectors were assumed to be expensive, at a cost of \$45{,}000/kg (2024 USD). The final set of solutions corresponding to the Pareto fronts generated by the eight agents is shown in Figure \ref{fig:pearlresultsnominal}. As expected, a clear trade-off emerges between $F_{\Delta h}$ and LCOE, particularly for extreme values of the latter (i.e., below \$7{,}000/MWh). Notably, as observed in \citep{seurin2025techno}, a large number of solutions satisfy $F_{\Delta h} \leq 1.47$, indicating that the nominal design was not optimized for this QOI. Although an arbitrary number of ``final'' designs could have been selected for human review by using techniques such as the entropy method \citep{kumar2016multi,seurin2024multiobjective}, for the sake of brevity we present only two representative designs in this paper. These correspond to the best-performing solutions identified for each objective.

Tables \ref{tab:inparmsnominal} and \ref{tab:nominaldesignobjectivesnominal} compare the best designs identified for each objective, the best design obtained through single-objective LCOE optimization (see \citep{seurin2025techno}), and the nominal design. Notably, the best solutions differ primarily in terms of $x_{ca}$ and $x_{pp}$, which are reduced to lower the $F_{\Delta h}$ (Solution 2), as discussed in Section \ref{sec:samplesdescription}. Additionally, $x_{mr}$ is consistently smaller than in the nominal case, reflecting reduced moderation and increased core reactivity. Moreover, yttrium hydride exhibits positive reactivity coefficients \citep{Ade02122022}, and so reducing its presence lowers the ITC for all solutions, as is consistent with observations in \citep{seurin2025techno}. Typically, a smaller $x_{mr}$ also decreases the SDM and $F_{\Delta h}$, favoring safety, while other mechanisms are employed to improve the fuel lifetime.

Furthermore, $x_{fh}$ is maximized in all cases to reduce the amount of expensive axial Be reflector material \citep{seurin2025techno} and potentially lower $F_{\Delta h}$. Consequently, $q^{''}_{\text{max}}$ remains low across all solutions, as is consistent with the correlation matrix in Figure \ref{fig:correlationmatrix}.

The LCOE for Solution 1 (\$4,992/MWh) is significantly lower than that of the best single-objective case (i.e., \$5,844/MWh), as driven by an extended fuel lifetime and higher burnup. However, the SDM for Solution 1 violates the imposed constraint. It is worth noting that the SDM values for HPMRs are substantially higher than those for light-water reactors and small modular reactors (e.g., $-1{,}200$ pcm \citep{halimi2024scale,seurin2024assessment} versus $-6{,}725$ pcm for the nominal case). Relaxing this constraint could improve economic performance, representing an area for future research.

The higher burnup in the least expensive solution reduces the fuel's contribution to LCOE, albeit marginally, as most costs stem from capital and O\&M expenses (see \citep{seurin2025techno}) and initial fuel inventory.

While these results provide valuable insights, further improvements will require additional data and surrogate model tuning. Each sample is computationally intensive ($>$48 hours)---comparable to the time required for the entire optimization to run. This motivates exploration of advanced strategies such as active learning \citep{nabila5771943active} or dynamic retraining to enhance surrogate accuracy---an active area of ongoing research.

\begin{table}[htp!]
\centering
\caption{Design input parameters for nominal cost scenario designs. Solution 1 is the best design from an LCOE perspective, while Solution 2 is best in terms of $F_{\Delta h}$.}
\begin{tabular}{lllll}
\hline
Symbol    &  Solution 1 & Solution 2 &    Solution (single-obj)     & Nominal\\ \hline
$x_{ca}$  & 86.00 & 35 & 91 & 90 \\
$x_{B10}$ & 0.20 & 77.8 & 53 & 95\\
$x_{fh}$  & 190 & 190 & 190 & 160\\
$x_{pp}$  & 1.94 & 2.31 & 2.20 & 2.3\\
$x_{e}$   & 0.199 & 0.199 & 0.199 & 0.197\\
$x_{cr}$  & 0.97 & 1.01 & 1.10 & 1\\
$x_{mr}$  & 0.743 & 0.716 & 0.75 & 0.825\\
\hline
\end{tabular}
\label{tab:inparmsnominal}
\end{table}

\begin{table}[htp!]
    \centering
     \caption{QoI results for the nominal cost scenario designs. The true values are given on the left of each slash, and the predicted ones on the right.}
    \begin{tabular}{lllll}
    \hline
    QoIs & Solution 1 & Solution 2 & Solution (single-obj) & Nominal \\
    \hline
      Lifetime   & 14.03/10.36  & 4.97/7.31 & 11.41/10.24 & 6.99/8.08\\
    SDM &-4830 /-7083 & -6805/-6786 &  -6708/-7417 & -6725/-6730\\
$F_{\Delta h}$& 1.333/1.467 & 1.317/1.315 & 1.41/1.40 & 1.469/1.474\\
ITC & -2.94 & -3.12 & -3.04 & -2.404 \\
$q^{''}_{max}$ & 0.0174/0.0187 & 0.0170/0.0167 & 0.016/0.016 & 0.0188/0.0192\\
Average Heat Flux & 0.0091 & 0.0088 & 0.00806 & 0.010536\\
Power Density & 1.876 & 1.76 & 1.612  & 2.105\\
Uranium U235 & 116.746 & 124.572 & 149.90 &103.44\\
Uranium Mass & 586.66 & 625.99 & 753.27 & 525.06\\
Burnup & 17.470 & 5.800 & 11.07 & 9.725\\
LCOE (FOAK) Estimated Cost (\$2024)& 4992/5523 & 6481/6941 & 5844/6190  & 10307 \\
\hline
    \end{tabular}
\label{tab:nominaldesignobjectivesnominal}
\end{table}

Note the $F_{\Delta h}$ found (see Solution 2 in Table \ref{tab:nominaldesignobjectivesnominal}) is the lowest compared with the best found in the scenarios studied in Section \ref{sec:applicationofbecostofgraphite} and \ref{sec:applicationofbecostofgraphiteeveruwhere}. Moreover, here the solution with the lowest LCOE found with multi-objective optimization (see Solution 1 in Table \ref{tab:nominaldesignobjectivesnominal}) surpasses that of the single-objective one with an LCOE amounting to 4992 against 5844 \$/MWh, respectively. This is due to the $x_{fh}$, which must be maximized to optimize both QoIs. In this case, two incentives drive the agents to look for a good solution in the low $F_{\Delta h}$ space. This is a manifestation of the physics-informed approaches highlighted in \citep{seurin2024physics}---by using a strategic combination of objectives and constraints in relation to the input space, multi-objective solutions may surpass the single-objective ones.

\subsubsection{Scenario Analysis: Inexpensive Axial Reflector with Expensive Control Drums}
\label{sec:applicationofbecostofgraphite}
\begin{figure}[htp!]
    \centering
    \includegraphics[width=0.8\linewidth]{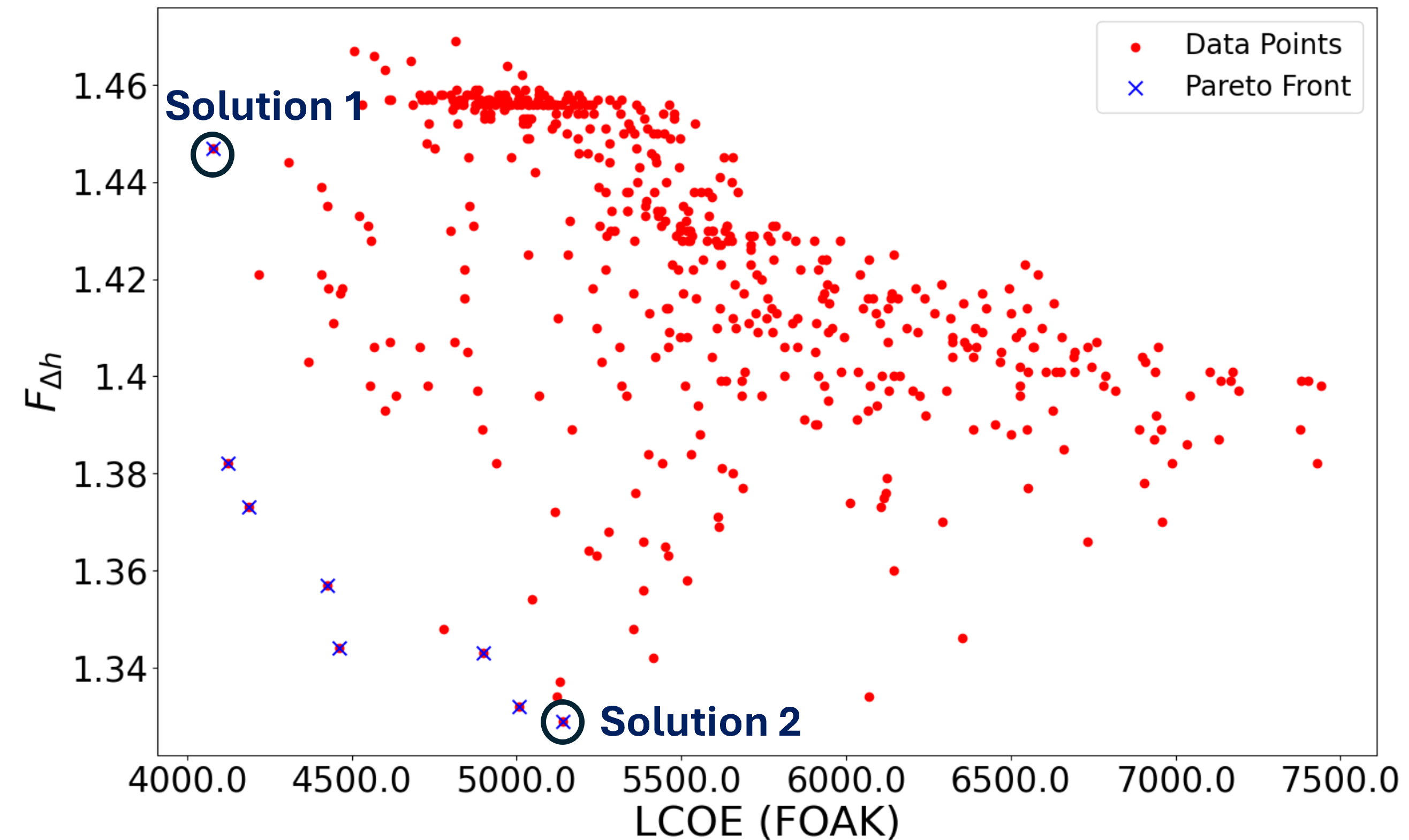}
    \caption{Pareto fronts from PEARL agents for the scenario with an inexpensive axial reflector. The two solutions corresponding to the minimum LCOE (i.e., Solution 1) and $F_{\Delta h}$ (i.e., Solution 2) are circled in blue.}
    \label{fig:pearlresultsgraphite}
\end{figure}

In this scenario, the axial reflector is assumed to be inexpensive, priced at the value of graphite (\$80/kg in 2022 USD), whereas the control drums are considered costly. The final set of solutions corresponding to the Pareto fronts identified by each of the eight agents is presented in Figure~\ref{fig:pearlresultsgraphite}.

\begin{table}[htp!]
\centering
\caption{Design input parameters for the scenario with inexpensive axial reflectors.}
\begin{tabular}{lllll}
\hline
Symbol    &  Solution 1 & Solution 2 &    Solution (single-obj)     & Nominal\\ \hline
$x_{ca}$  & 106.7 & 56.6 & 96.0 & 90 \\
$x_{B10}$ & 77.7 & 50.4  & 61.5 & 95\\
$x_{fh}$  & 178.2 & 190  & 148.36 & 160\\
$x_{pp}$  & 1.94 & 2.14  & 1.94 & 2.3\\
$x_{e}$   & 0.186 & 0.199& 0.199 & 0.197\\
$x_{cr}$  & 0.97 & 1.07  & 0.97  & 1\\
$x_{mr}$  & 0.80 & 0.742 & 0.689 & 0.825\\
\hline
\end{tabular}
\end{table}

\begin{table}[htp!]
    \centering
     \caption{QoI results for the scenario with inexpensive axial reflector designs.}
    \begin{tabular}{lllll}
    \hline
    QoIs & Solution 1 & Solution 2 & Solution (single-obj) & Nominal \\
    \hline
      Lifetime   & 10.72/8.653 & 9.00/8.78 & 9.61/8.71 & 6.99/8.08\\
    SDM & -7338/-6951 & -5659/-7627 & -7952/-7473 & -6725/-6730\\
$F_{\Delta h}$& 1.455/1.447 & 1.37/1.329 & 1.373/1.343 & 1.469/1.474\\
ITC & -2.74 & -2.93 & -3.00 & -2.404 \\
$q^{''}_{max}$ & 0.0188/0.0188 & 0.0164/0.0159 &  0.0214/0.0214  & 0.0188/0.0192\\
Average Heat Flux & 0.00974 & 0.0083 & 0.0117 & 0.010536\\
Power Density & 2.008 & 1.551 & 2.412 & 2.105\\
Uranium U235 & 102.515 & 142.084 & 91.160 &103.44\\
Uranium Mass & 550.241 & 713.99 & 458.09 & 525.06\\
Burnup & 14.23 & 9.21 & 15.33 & 9.725\\
LCOE (FOAK) Estimated Cost (\$2024)& 4052/4838 & 4906/4924 & 3937/3959 & 5079\\
\hline
    \end{tabular}
\label{tab:nominaldesignobjectivespartialgraphite}
\end{table}
In Figure~\ref{fig:pearlresultsgraphite}, we observe clear trade-offs between the LCOE and $F_{\Delta h}$, particularly at lower values of the former. Tables~\ref{tab:inparmsnominal} and \ref{tab:nominaldesignobjectivesnominal} compare the best designs (with respect to each objective), the optimal design obtained from single-objective LCOE optimization (see \cite{}), and the nominal design. Solution 1 and the best single-objective design differ primarily in terms of $x_{fh}$ and $x_{mr}$. In the latter, the axial reflector volume is maximized to reduce the control drums' contribution to LCOE by substituting expensive TRISO fuel with cost-effective graphite reflector material. This substitution also enhances reactivity, enabling a reduction in $x_{mr}$ and consequently lowering $F_{\Delta h}$. Conversely, Solution 1, which was comparable to the single-objective cost result within uncertainty bounds---exhibits a higher $x_{fh}$. To compensate for the reduced reactivity, $x_{mr}$ must be increased, in turn raising $F_{\Delta h}$ (1.455 versus 1.373 for Solution 1 and the single-objective solution, respectively). An additional benefit of the increased $x_{fh}$ is a reduction in $q^{''}_{\text{max}}$. The single-objective design, characterized by the lowest $x_{fh}$, therefore results in a significantly higher average heat flux and the highest $q^{''}_{\text{max}}$, as is consistent with the correlation trends shown in Figure~\ref{fig:correlationmatrix}.


Finally, as discussed in Section~\ref{sec:applicationofbecostofnominal}, while the surrogate models used in this study could be further refined, they nonetheless provide valuable insights into the trade-offs between cost and safety in HPMR design.

\subsubsection{Scenario Analysis: Inexpensive Axial and Drum Reflectors}
\label{sec:applicationofbecostofgraphiteeveruwhere}
\begin{figure}[htp!]
    \centering
    \includegraphics[width=0.8\linewidth]{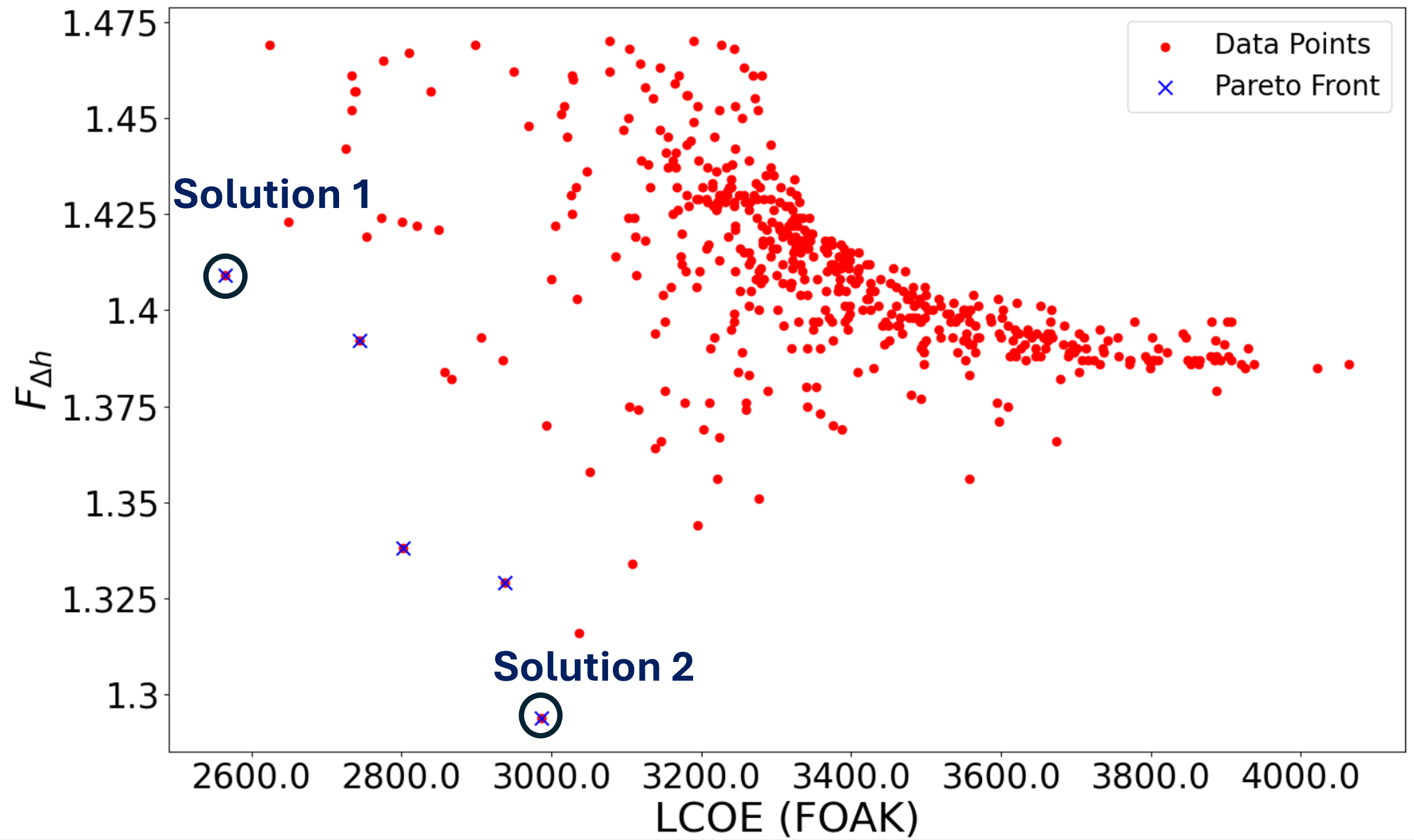}
    \caption{Pareto fronts from PEARL agents for the scenario with inexpensive axial and drum reflectors. The two solutions corresponding to the minimum LCOE (i.e., Solution 1) and $F_{\Delta h}$ (i.e., Solution 2) are circled in blue.}
    \label{fig:pearlresultsfullgraphite}
\end{figure}

In this scenario, both the axial and drum reflectors are assumed to be inexpensive, with Be priced at \$80/kg (2023 USD). The final set of solutions corresponding to the Pareto fronts identified by each of the eight agents is shown in Figure~\ref{fig:pearlresultsfullgraphite}. As observed previously, a clear trade-off exists between LCOE and $F_{\Delta h}$, particularly at lower LCOE values. Notably, the constraint $F_{\Delta h} \leq 1.47$ is readily satisfied in this case, but the extent of the Pareto front is reduced, owing to the lesser contributions of the Be reflectors.

\begin{figure}
    \centering
    \includegraphics[width=0.5\linewidth]{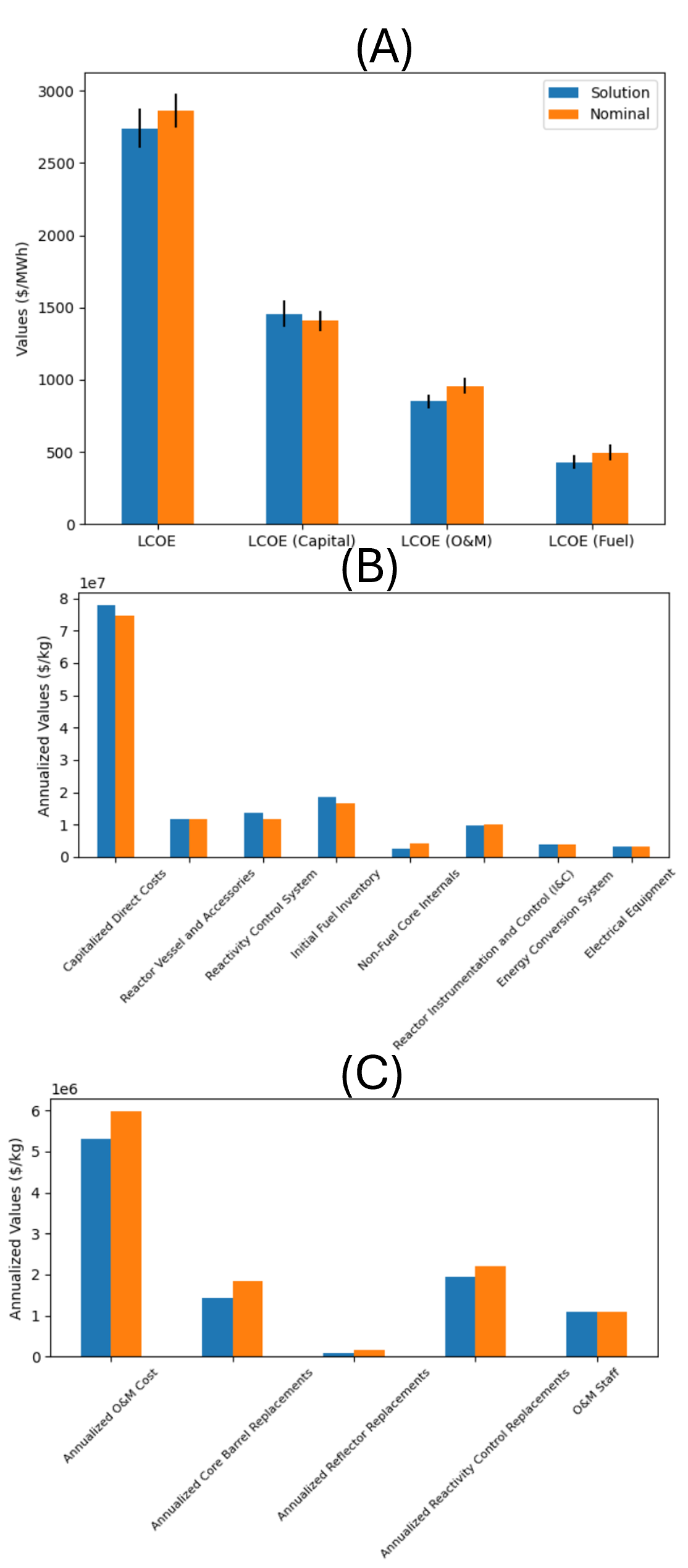}
    \caption{LCOE breakdown across different scenarios: (A) LCOE breakdown for the nominal design, (B) LCOE breakdown for the scenario with an inexpensive axial reflector, and (C) LCOE breakdown for the scenario with inexpensive axial and drum reflectors.}
    \label{fig:breakdownoflcoe}
\end{figure}

Despite the lower drum reflector costs, the cost-minimization strategy remains consistent with that described in Section~\ref{sec:applicationofbecostofgraphite}. Specifically, reducing $x_{fh}$ to replace expensive TRISO fuel with graphite and minimizing the control drum contribution to LCOE continues to be favored, as highlighted in Table~\ref{tab:inparmsfullgraphite}. Although beryllium is now inexpensive, the boron carbide ($\mathrm{B_4C}$) used in the drums remains costly at \$14,268/kg (2023 USD). This is confirmed by Figures~\ref{fig:breakdownoflcoe}(B) and~\ref{fig:breakdownoflcoe}(C), which show that the ``Reactivity Control System'' and ``Annualized Reactivity Control Replacements'' still contribute significantly to the overall LCOE.

Additionally, the optimized designs in this scenario exhibit higher fuel burnup, reducing the fuel cycle's contribution to LCOE in comparison to the nominal design (see Figure~\ref{fig:breakdownoflcoe}(A)). Accordingly, two primary strategies emerge:
\begin{enumerate}
    \item Replace expensive TRISO fuel with graphite reflector material and reduce reliance on control drums.
    \item Increase fuel burnup to lower the fuel cycle's contribution to LCOE.
\end{enumerate}

Furthermore, while the lowest-LCOE solution in the previous scenario (obtained via single-objective optimization (see Table~\ref{tab:nominaldesignobjectivespartialgraphite}) exhibited the highest $q^{''}_{\text{max}}$, the lowest-LCOE solution in this scenario achieves a comparatively lower $q^{''}_{\text{max}}$ (see Table~\ref{tab:nominaldesignobjectivesfullgraphite}). This improvement is attributed to the higher $x_{fh}$, which was chosen due to the reduced cost of the drum reflectors (with only the $\mathrm{B_4C}$ component remaining expensive). This configuration enables better thermal performance while satisfying the $q^{''}_{\text{max}}$ constraint.

The primary distinction between the multi-objective Solution 1 and the single-objective solution in this scenario lies in the safety factor $F_{\Delta h}$, which is influenced by differences in $x_{pp}$ and $x_{mr}$. In the single-objective solution, a higher $x_{pp}$ supports an extended fuel lifetime, with a lower $x_{mr}$ and hence a lower ITC, but also results in a higher $F_{\Delta h}$ (1.45).

\begin{table}[htp!]
\centering
\caption{Design input parameters for the scenario with inexpensive axial and drum reflectors.}
\begin{tabular}{lllll}
\hline
Symbol    &  Solution 1 & Solution 2 &    Solution (single-obj)     & Nominal\\ \hline
$x_{ca}$  & 100.82 & 35 & 98 & 90 \\
$x_{B10}$ & 20 & 20 & 95 & 95\\
$x_{fh}$  & 172 & 190 & 158 & 160\\
$x_{pp}$  & 1.94 & 2.61 & 2.78 & 2.3\\
$x_{e}$   & 0.197 & 0.199 & 0.197 & 0.197\\
$x_{cr}$  & 0.908 & 0.948 & 1.063 & 1\\
$x_{mr}$  & 0.733 & 0.702 & 0.522 & 0.825\\
\hline
\end{tabular}
\label{tab:inparmsfullgraphite}
\end{table}

\begin{table}[htp!]
    \centering
    \caption{QoI results for the scenario with inexpensive axial and drum reflector designs.}
    \begin{tabular}{lllll}
    \hline
    QoIs & Solution 1 & Solution 2 & Solution (single-obj) & Nominal \\
    \hline
      Lifetime   & 6.58/8.78 & 4.42/8.83  & 11.42/9.87 & 6.99/8.08\\
                
    SDM & -7956/-7264 & -6277/-7862 & -6557/-6745 & -6725/-6730\\
                
$F_{\Delta h}$& 1.387/1.329 & 1.365/1.294 & 1.455/1.476 & 1.469/1.474\\
$q^{''}_{max}$ & 0.0192/0.0159 & 0.0182/0.0174 & 0.0182/0.0186 & 0.0188/0.0192\\
ITC & -2.91 & -2.88 & -3.63 & -2.404 \\
Average Heat Flux & 0.0108 & 0.00935 & 0.0100  & 0.010536\\
Power Density & 2.379 & 1.973 & 1.881 & 2.105\\
Uranium U235 & 91.773 & 111.4 & 115.82 & 103.44\\
Uranium Mass & 465.63 & 559.80 & 586.83 & 525.06\\
Burnup & 10.32 & 5.768 & 14.216 & 9.725\\
LCOE (FOAK) Estimated Cost (\$2024)& 2786/2621 & 3317/2941 & 2630/2754 & 2857\\
\hline
    \end{tabular}
\label{tab:nominaldesignobjectivesfullgraphite}
\end{table}

Lastly, the significant discrepancies observed in fuel lifetime once again highlight the critical need for improving the surrogate modeling approach.

\subsection{Summary of the Studies Conducted}
\label{sec:summary}

In this work, we applied MOO to explore the trade-offs between cost---as quantified via the LCOE---and safety---represented by the rod-integrated peaking factor $F_{\Delta h}$---while satisfying a set of operational and safety constraints. Three distinct scenarios were investigated:

\begin{enumerate}
    \item In Section~\ref{sec:applicationofbecostofnominal}, we assumed Be in the axial reflector, which is expensive but would allow the most compact core. The optimal strategy for minimizing LCOE involved reducing the amount of Be in the reflector, thereby maximizing $x_{fh}$. Interestingly, this strategy also aligned with minimizing $F_{\Delta h}$, resulting in a MOO solution (Solution 1 in Table~\ref{tab:nominaldesignobjectivesnominal}) that outperformed both the single-objective and nominal designs. This outcome exemplifies a form of physics-informed RL in which the alignment of objectives allows MOO to yield superior cost performance \citep{seurin2024physics}.
    \item As covered in \ref{sec:applicationofbecostofgraphite}, we considered a scenario with a low-cost axial reflector, which would typically be graphite, allowing some LCOE reduction to the cost of a less compact core. Here, the control drums emerged as the primary cost driver. The optimization strategy involved reducing $x_{fh}$ to substitute expensive TRISO fuel with graphite and to lessen the drums' contribution to LCOE. However, $x_{fh}$ could not be minimized indefinitely, due to its indirect influence on $F_{\Delta h}$ and $q^{''}_{\text{max}}$. Additionally, the fuel burnup was further increased to more significantly reduce the fuel cycle's contribution to LCOE than was the case in the previous scenario, also impacting the initial fuel inventory and associated capital costs.
     \item As covered in \ref{sec:applicationofbecostofgraphiteeveruwhere}, we assumed both the axial and drum reflectors to be inexpensive. This would hold typically for a fully LCOE-optimized core. Although the cost drivers remained similar to those in the previous scenario, their impact was reduced, as only the $\mathrm{B_4C}$ component in the drums remained expensive. Consequently, the strategy again favored reducing $x_{fh}$ and further increasing the fuel burnup so as to minimize the fuel cycle's contribution to LCOE, this time even more aggressively than in the preceding scenarios.
\end{enumerate}

Across all three scenarios, several consistent design trends emerged. The control drum coating angle $x_{ca}$ was consistently minimized to reduce $F_{\Delta h}$. Minimizing $x_{mr}$ and $x_{pp}$ also proved beneficial for lowering $F_{\Delta h}$, though these reductions negatively impacted the fuel lifetime, necessitating trade-offs between safety and operational longevity. Additionally, maximizing $x_{fh}$ was a recurring strategy across all scenarios to reduce $F_{\Delta h}$.

Finally, increasing the fuel burnup consistently improved LCOE by reducing the fuel cycle cost and the initial fuel inventory, both of which contribute to the capital expenditures. However, the extent of this benefit varied across the different scenarios, being less pronounced in the first and second cases, due to the dominant influence of the axial reflector and drum costs.
\section{Conclusion}
\label{sec:conc}

$\mu$Rs are nuclear reactors that generate less than 20~MW$_\text{th}$ of power, causing them to be classified as Category B reactors as defined by the U.S.~Department of Energy. Consequently, they fall under Category 2 hazards as per 10 CFR 830 \citep{naranjo2024assessment}. Designed for factory assembly, transportability (via truck, rail, or cargo), simplified installation, and self-regulation, $\mu$Rs are particularly well-suited for deployment in remote, unmanned environments in which reliance on costly fossil fuels persists (e.g., deep-sea operations, industrial mining, and U.S.~military bases). Despite their operational advantages, $\mu$Rs suffer from diseconomies of scale and must rely on economies of multiples to achieve market competitiveness. Within this landscape, HPMRs distinguish themselves by employing passive HP cooling, eliminating the need for circulation pumps and auxiliary systems and thus enabling highly compact designs. However, HPMRs often face economic disadvantages in comparison to alternatives such as gas- or lead-cooled $\mu$Rs \citep{hanna2025bottom}. Therefore, optimizing the economic performance of HPMRs is essential, particularly given the complex interplay between cost and safety---an area that remains insufficiently understood. Advances in AI-based optimization methods offer promising avenues for exploring these trade-offs. In this study, we employed a multi-objective RL approach (i.e., PEARL) to jointly optimize LCOE and $F_{\Delta h}$ while satisfying all constraints on the $q^{''}_{\text{max}}$, fuel lifetime, and SDM. Our results confirm the value of true multi-objective optimization in terms of elucidating the trade-offs inherent in HPMR design decisions. We investigated three cost scenarios:
\begin{enumerate}
    \item \textbf{Expensive axial and drum reflectors (Section~\ref{sec:applicationofbecostofnominal}):} Maximizing $x_{fh}$ proved beneficial for both LCOE and $F_{\Delta h}$, reducing the use of costly axial reflector material and flattening the power profile. This alignment of objectives enabled multi-objective optimization to outperform single-objective approaches, exemplifying physics-informed RL \citep{seurin2024physics}.
    
    \item \textbf{Inexpensive axial reflector (\ref{sec:applicationofbecostofgraphite}):} With the axial reflector cost reduced, control drums became the dominant cost contributor. Strategies focused on reducing $x_{fh}$ to substitute expensive TRISO fuel with graphite and to minimize drum reliance. Fuel burnup was increased to reduce both fuel cycle costs and the initial fuel inventory, achieving an impact greater than in Scenario 1.
    
    \item \textbf{Inexpensive axial and drum reflectors (\ref{sec:applicationofbecostofgraphiteeveruwhere}):} With both reflector components being inexpensive (except for $\mathrm{B_4C}$ in the drums), strategies similar to those in Scenario 2 were employed, but with more aggressive burnup increases to further reduce fuel cycle costs.
\end{enumerate}
Across all scenarios, several consistent design strategies emerged:
\begin{itemize}
    \item Minimizing $x_{ca}$ (i.e., the drum coating angle) to reduce $F_{\Delta h}$
    \item Maximizing $x_{fh}$ (i.e., the active fuel height) to improve both cost and safety performance
    \item Balancing $x_{pp}$ (i.e., the pin pitch) and $x_{mr}$ (i.e., the moderator radius) to manage trade-offs between the peaking factor and fuel lifetime
    \item Increasing fuel burnup to reduce both the fuel cycle portion of LCOE and the initial fuel inventory, thereby lowering capital costs.
\end{itemize}
Notably, $q^{''}_{\text{max}}$ (i.e., the peak heat flux) remained close to nominal values across all scenarios, even when relaxed to 0.025 (from 0.0188 in the nominal design), indicating that it does not substantially influence cost in HPMRs.

While achieving $F_{\Delta h}$ (i.e., the rod integrated peaking factor) lesser than $1.47$ was relatively straightforward, minimizing LCOE proved more scenario-dependent. In Scenario 1, maximizing $x_{fh}$ improved both objectives, enabling the single-objective optimizer to find a solution comparable to the multi-objective one. In Scenarios 2 and 3, minimizing control drum contributions---due to the cost of Be and $\mathrm{B_4C}$---was key, with Scenario 2 showing a stronger cost sensitivity to drum materials.

Despite the success of the PEARL-based optimization, discrepancies between the surrogate model predictions and full-order simulations were observed, particularly as regards fuel lifetime and SDM. While these discrepancies did not significantly affect the single-objective optimization results \citep{seurin2025techno}, they may impact multi-objective optimization, in which multiple constraints interact. In our study, SDM and lifetime constraints were generally satisfied, with minor violations in Solution 1 and Solution 2 for the first scenario (Table~\ref{tab:nominaldesignobjectivesnominal}). Nevertheless, the surrogate models effectively guided the optimizer toward promising design regions, supporting the overall optimization process. This highlights the importance of continued development in surrogate modeling, especially for computationally intensive systems. Future work will explore active learning and dynamic retraining to improve surrogate accuracy and robustness.

In conclusion, the PEARL algorithm demonstrates strong potential for multi-objective optimization in HPMR design, enabling nuanced exploration of cost and safety trade-offs. Even in cases with surrogate discrepancies, the resulting designs met most operational constraints and revealed valuable strategies for improving economic competitiveness without compromising safety. Further research into constraint relaxation and surrogate refinement is warranted in order to unlock additional cost-saving opportunities.


\section*{Credit Authorship Contribution Statement}

 \textbf{Paul Seurin:} Conceptualization, Methodology, Software, Writing – original draft preparation, Funding acquisition, Data curation, Formal analysis \& investigation, Visualization. \textbf{Dean Price:} Conceptualization, Methodology, Software, Writing – review \& editing, Data curation, Funding acquisition.
 
\section*{Acknowledgment}
This work was supported through the Idaho National Laboratory (INL) Laboratory Directed Research \& Development (LDRD) program under U.S.~Department of Energy (DOE) Idaho Operations Office contract no.~DE-AC07-05ID14517. The authors would also like to thank Dr.Sonali Roy and Dr Botros Hanna for the technical review of the paper, and John Shaver for proof-reading the manuscript.

\section*{Data Availibility}
The data and codes will, by the time of this publication, be made available upon reasonable demand and in an open-source package hosted on the \href{https://github.com/IdahoLabResearch}{IdahoLabResearch GitHub repository}.

\bibliographystyle{elsarticle-num-names} 
\bibliography{refs.bib}

\end{document}